\documentclass[letterpaper, 10 pt, conference]{ieeeconf}  % Comment this line out if you need a4paper
\usepackage[utf8]{inputenc}

\IEEEoverridecommandlockouts                              % This command is only needed if 
\usepackage{graphics} % for pdf, bitmapped graphics files
\usepackage{epsfig} % for postscript graphics files
\usepackage{times} % assumes new font selection scheme installed
\usepackage{amsmath} % assumes amsmath package installed
\usepackage{amssymb}  % assumes amsmath package installed
\usepackage{hyperref}

\usepackage{ulem}       %% 删除线
\usepackage{subfig}     %% 图像拼接
\usepackage{booktabs}   %% 使用三线表：toprule，midrule，bottomrule
\usepackage{multirow}   %% 表格中支持跨行
\usepackage{cleveref}
\usepackage{tabularx}   %% 表格
\usepackage[table]{xcolor}     %% 文字上色
\usepackage[utf8]{inputenc}
\newlength\savewidth

\newcommand{\etal}{\textit{et al.}}
\newcommand{\ie}{\textit{i.e.}}
\title{\LARGE \bf
MRASfM: Multi-Camera Reconstruction and Aggregation through Structure-from-Motion in Driving Scenes}
\author{Lingfeng Xuan\textsuperscript{\dag1}, Chang Nie\textsuperscript{\dag1} Yiqing Xu\textsuperscript{2}, Zhe Liu\textsuperscript{1}, Yanzi Miao\textsuperscript{2}, and Hesheng Wang\textsuperscript{1} % <-this % stops a space
\thanks{\textsuperscript{\dag}The first two authors contributed equally.}
\thanks{This work was supported in part by the Natural Science Foundation of China under Grant 62225309, U24A20278, 62361166632 and U21A20480. (Corresponding Author: Hesheng Wang)
}% <-this % stops a space
\thanks{\textsuperscript{1}Department of Automation,
Key Laboratory of System Control and Information Processing of Ministry of
Education, Key Laboratory of Marine Intelligent Equipment and System of
Ministry of Education, Shanghai Engineering Research Center of Intelligent
Control and Management, Shanghai Jiao Tong University, Shanghai 200240,
China. }
\thanks{\textsuperscript{2}The Advanced Robotics Research Center, Artificial Intelligence Research Institute and School of Information and Control Engineering, China University of Mining and Technology, Xuzhou, 221116, China.}
}

\begin{document}
\maketitle
\thispagestyle{empty}
\pagestyle{empty}
	
%%%%%%%%%%%%%%%%%%%%%%%%%%%%%%%%%%%%%%%%%%%%%%%%%%%%%%%%%%%%%%%%%%%%%%%%%%%%%%%%
\begin{abstract}

% This paper introduces a 3D point cloud sequence learning model based on inconsistent spatio-temporal propagation for deep LiDAR odometry, termed DSLO.
Structure from Motion (SfM) estimates camera poses and reconstructs point clouds, forming a foundation for various tasks. However, applying SfM to driving scenes captured by multi-camera systems presents significant difficulties, including unreliable pose estimation, excessive outliers in road surface reconstruction, and low reconstruction efficiency. To address these limitations, we propose a Multi-camera Reconstruction and Aggregation Structure-from-Motion (MRASfM) framework specifically designed for driving scenes. MRASfM enhances the reliability of camera pose estimation by leveraging the fixed spatial relationships within the multi-camera system during the registration process. To improve the quality of road surface reconstruction, our framework employs a plane model to effectively remove erroneous points from the triangulated road surface. Moreover, treating the multi-camera set as a single unit in Bundle Adjustment (BA) helps reduce optimization variables to boost efficiency. In addition, MRASfM achieves multi-scene aggregation through scene association and assembly modules in a coarse-to-fine fashion. We deployed multi-camera systems on actual vehicles to validate the generalizability of MRASfM across various scenes and its robustness in challenging conditions through real-world applications. Furthermore, large-scale validation results on public datasets show the state-of-the-art performance of MRASfM, achieving 0.124 absolute pose error on the nuScenes dataset. 
%Our implementation will be available at \url{https://github.com/IRMVLab/MRASfM}.

% Spatial features are encoded and reused to reduce computational overhead. Sequential pose initialization leverages LiDAR's high-frequency sampling for pose estimation. Gated hierarchical refinement refines poses from coarse to fine using gate estimations. Temporal feature propagation incorporates historical motion information and addresses spatial inconsistencies between frames. Experiments on the KITTI and Argoverse datasets show DSLO outperforms state-of-the-art methods with at least a 15.67\% improvement in RTE, a 12.64\% improvement in RRE, and a 34.69\% reduction in runtime. Our implementation will be available at \url{https://github.com/IRMVLab/DSLO}.

\end{abstract}

%%%%%%%%%%%%%%%%%%%%%%%%%%%%%%%%%%%%%%%%%%%%%%%%%%%%%%%%%%%%%%%%%%%%%%%%%%%%%%%%
\section{Introduction}
\label{sec:intro}
Image-based driving scene reconstruction is commonly achieved through visual Simultaneous Localization and Mapping (vSLAM) and Structure from Motion (SfM).  
However, for critical downstream tasks like high-definition (HD) mapping construction and novel view synthesis \cite{xie2023mv,li20243d,zhao2024hybridocc}, the limitations of vSLAM become apparent.  
Specifically, its reliance on incremental, locally optimized estimations impedes global refinement, leading to accumulated drift and reduced accuracy. 
In contrast, SfM leverages batch processing and global Bundle Adjustment (BA), thereby yielding more accurate scene point clouds and camera trajectories. 
%This offline nature allows SfM to circumvent the accuracy compromises inherent in real-time vSLAM, making it demonstrably superior for achieving the precision and consistency demanded by those tasks.
This offline nature allows SfM to overcome the accuracy trade-offs inherent in real-time vSLAM, making it more suitable for tasks requiring high precision and consistency.
However, the application of SfM to driving scenes captured by multi-camera systems encounters notable obstacles, including unreliable pose estimation, excessive outliers in road surface reconstruction, and low reconstruction efficiency.

The reliability of camera pose estimation in driving scenes is inherently challenged by the characteristics of these environments.
Traditional SfM typically captures images around the object, while it is difficult to implement in driving scenes. 
Moreover, repetitive patterns and dynamic objects also decrease the quality of correspondence search, making pose estimation unreliable in driving scenes.

Road surface reconstruction is another significant challenge for SfM in driving scenes, which is crucial for downstream tasks \cite{wu2024emie,dai2024high}.
The edges of vehicle shadows are often detected as feature points.
When the vehicle shadows move, the corresponding feature points become dynamic points, introducing noise into road surface reconstruction.
Moreover, the lack of texture also causes outliers. % 点明噪声很多

\begin{figure}[t]  
  \centering  
  \includegraphics[width=0.85\linewidth]{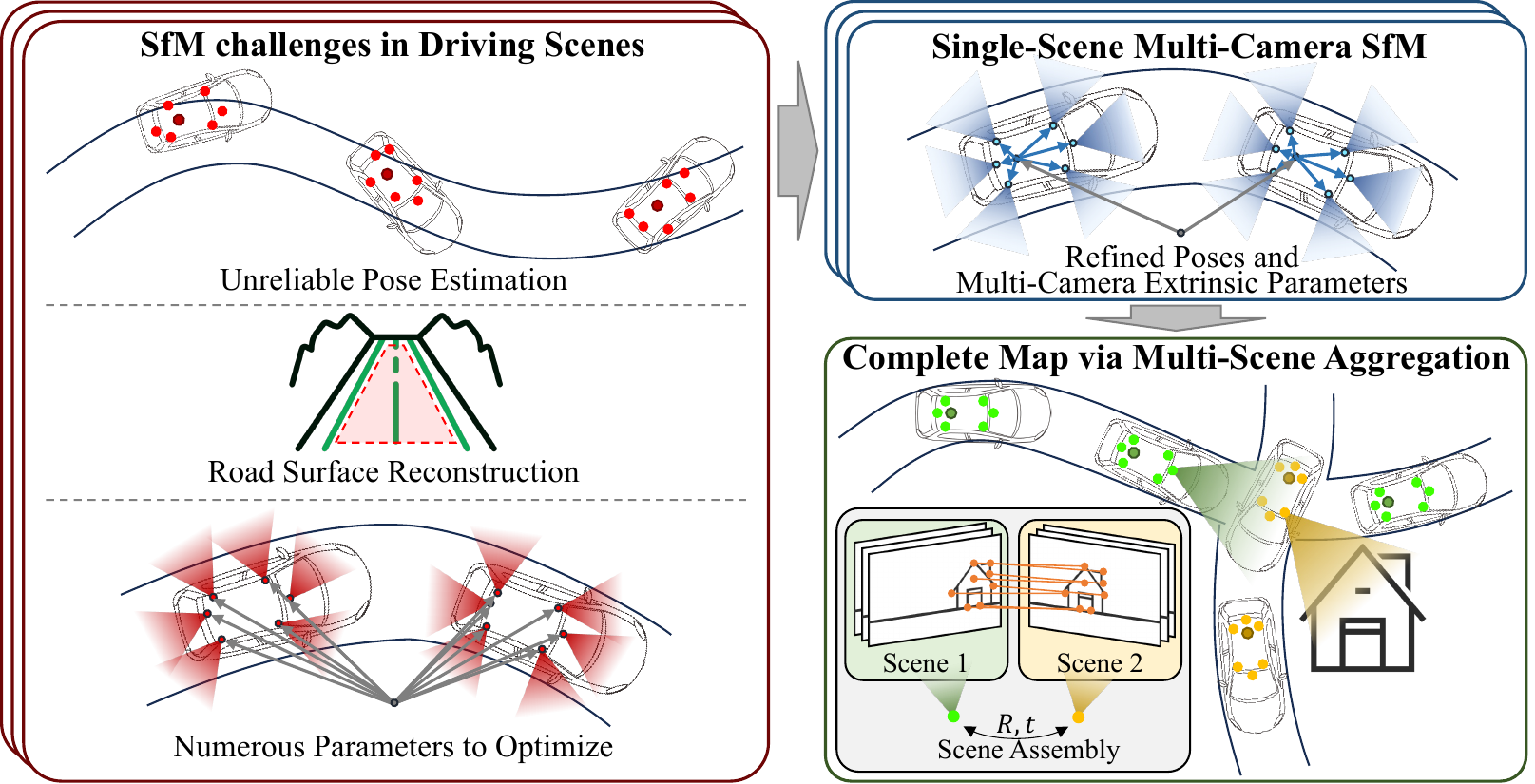} % 确保取消注释并使用实际图形文件  
  \vspace{-0.2cm}
  \caption{\textbf{The idea of MRASfM.} To address the challenges of SfM in driving scenes, MRASfM incorporates camera set priors and semantic information into the reconstruction framework, achieving efficient and robust scene reconstruction. Through multi-scene aggregation, fragmented scenes are integrated into a complete and consistent map. }  
  \label{fig:headpic}  
  \vspace{-0.6cm}
\end{figure}   
%For driving scene reconstruction, previous SfM-based works focus on single-camera systems  \cite{song2015high,song2015joint}. 
Efficiency remains a critical bottleneck in SfM, particularly in multi-camera systems.
While multi-camera setups offer a broader field of view (FOV) and the ability to capture more comprehensive environmental data, they also introduce complexities.
Unlike single-camera systems, which may fail entirely when obstructed, multi-camera systems can leverage alternative perspectives to maintain functionality. 
However, more cameras also bring more poses to be optimized in BA, thereby diminishing reconstruction efficiency.

To overcome these challenges, we propose a Multi-camera Reconstruction and Aggregation Structure-from-Motion (MRASfM) framework for driving scenes, as shown in Fig. \ref{fig:headpic}.
For unreliable pose estimation, our method incorporates a learning-based feature extractor to improve feature extraction. 
Moreover, prior camera poses are utilized to identify image pairs with significant visual overlap for matching, thereby enhancing the efficiency and robustness of matching.
During image registration, we initially estimate the camera poses of images with rich correspondences.
Leveraging the fixed spatial relationships of the multi-camera system, we can then robustly register even those images that are partially occluded. 
To mitigate excessive outliers in road surface reconstruction, we apply plane fitting to filter erroneous points and enhance the consistency of the reconstructed road surface. 
To improve the reconstruction efficiency, we enhance BA by treating the camera set as a unified unit.
In this way, optimization results are more consistent, accelerating the convergence of subsequent BA.
In addition, the reduction of optimization variables also boosts the efficiency of reconstruction.

In practical engineering, scenes are often captured over short durations with limited overlap \cite{mei2024rome,zhang2024vision}. 
However, generating a comprehensive scene understanding requires the integration of these fragmented reconstructions into a unified map.  
Traditional SfM aggregation modules typically require shared images between different segments \cite{cohen2015merging,chen2023adasfm}, which are often unavailable in practical applications.
%However, in practical scenes, shared images are often unavailable.
For such disjointed reconstructed scenes, we first use reconstruction results and GNSS to associate nearby scenes. % and assemble them coarsely.
% 从场景角度讲assembly， coarse bind
Then, images with large visual overlap are identified for matching and fine assembly.
Finally, an SfM-based optimization refines the transformation matrix between associated scenes for fine assembly.

In summary, the main contributions of our work are:
\begin{itemize}  
    \item 
    We propose MRASfM, a novel framework for multi-camera reconstruction, which overcomes the limitations of traditional SfM in pose estimation robustness and computational efficiency in driving scenes.
    % MRASfM can also calibrate cameras online during reconstruction.
    %Additionally, it can perform multi-scene in a coarse-to-fine fashion.
    \item 
    MRASfM introduces a camera set registration module, enhancing pose estimation robustness.
    Moreover, the camera set BA module greatly improves efficiency and achieves online calibration. 
    For fragmented scenes, the multi-scene aggregation module seamlessly binds them together in a coarse-to-fine fashion.
    %Transformation matrices between scenes are treated as optimization variables and iterative optimization is conducted among images with overlapping fields of view.
    \item 
     Experimental results on real-world applications demonstrate the generalizability of MRASfM across various environments and its robustness under challenging conditions.
    Moreover, experiments on public datasets highlight the state-of-the-art performance of MRASfM.
\end{itemize}  

\begin{figure*}[t]  
  \centering  
  \includegraphics[width=0.9\textwidth]{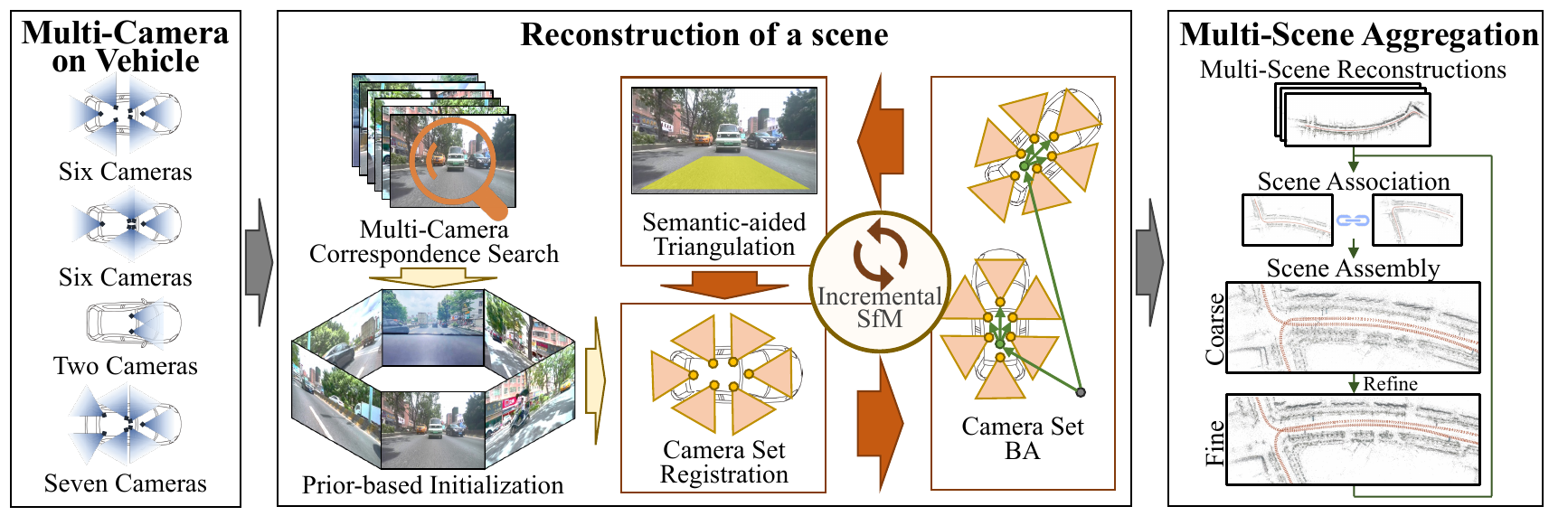} 
  \vspace{-0.2cm}
  \caption{\textbf{The pipeline of MRASfM.} MRASfM takes multi-camera images, semantic information, rough trajectories, and calibrations as input.  
  When reconstructing a scene, MRASfM first utilizes prior information to perform correspondence search and initialization (see \cref{subsec:Correspondence Search and Initialization}).
  In camera set registration, MRASfM robustly registers new images with rigid unit constraints (see \cref{subsec:RIG_Registratio}).
  During triangulation, MRASfM improves road surface reconstruction quality using semantic information for fewer outliers (see \cref{subsec:Semantic-assisted Triangulation}).
  In camera set BA, MRASfM refines the reconstruction using rigid units as optimization variables, enhancing efficiency and robustness (see \cref{subsec:RIG BA}).
  When aggregating multiple scenes, MRASfM assembles nearby scenes with SfM in a coarse-to-fine fashion (see \cref{subsec:Multi-scene Aggregation}).}
  \label{fig:PIPELINE}  
  \vspace{-0.6cm}
\end{figure*}

%%%%%%%%%%%%%%%%%%%%%%%%%%%%%%%%%%↑
\vspace{-0.2cm}
\section{Related Work}
\label{sec:formatting}
\vspace{-0.1cm}
\subsection{Multi-camera-Based Reconstruction}
\vspace{-0.1cm}
Multi-camera systems offer enhanced perception capabilities for driving scene reconstruction.
Traditional vSLAM-based multi-camera reconstruction methods typically require precise internal relative poses and intrinsic parameters of the cameras as input. 
Systems like BAMF-SLAM \cite{10160905} and MAVIS \cite{10609982} exemplify this reliance, requiring pre-calibration of their multi-camera systems for efficient localization and mapping. 
Moreover, these vSLAM methods struggle to be compatible with different multi-camera configurations, restricting their flexibility in practical deployments.

In the field of SfM, COLMAP \cite{schonberger2016structure} ignores camera set priors during incremental reconstruction. Instead, it applies these priors during Rigid Bundle Adjustment (RIG BA) after the reconstruction is complete. 
However, RIG BA is unable to rectify errors as they accumulate during the incremental reconstruction.  
MMA \cite{cui2022mma}, a global SfM method, seeks to improve accuracy and robustness by explicitly leveraging fixed relative camera poses. 
Despite its global nature, MMA is sensitive to feature match outliers and often degenerates when the estimated relative translations are collinear.
MGSfM \cite{tao2025mgsfm} achieves effective utilization of multi-camera constraints through decoupled rotation averaging and  hybrid translation averaging. 
Chen \etal \cite{chen2024camav2} presented an efficient odometry-guided SfM approach that directly uses odometry poses and camera extrinsics for image registration.  This method, however, demands highly precise extrinsic calibration and accurate odometry measurements, requirements that can be challenging to consistently meet in real-world driving scenes. 
MCSfM \cite{cui2023mcsfm} is a multi-camera-based incremental SfM system that does not require prior internal relative poses and intrinsic parameters. 
It allows for simultaneous camera set calibration during the reconstruction process. 
% 缺少细节 我们将xxx作为注册和优化的基本单元，提高效率和鲁棒性
Inspired by these works, we treat the camera set as the fundamental unit for registration and refinement, thus effectively integrating multi-camera systems into SfM and achieving efficient and robust reconstruction.

\vspace{-0.2cm}
\subsection{Incremental SfM}
\vspace{-0.2cm}
Incremental SfM \cite{schonberger2016structure,ozyecsil2017survey,wu2013towards} begins by establishing connections between images through correspondence searches. Next, an initial model is built using select seed images. 
Then, SfM iteratively performs image registration, triangulation, refinement, and outlier removal until all images are integrated into the system.

In correspondence search, traditional SIFT feature descriptors \cite{ng2003sift} generate limited feature points in low-texture images. 
In contrast, learning-based methods \cite{detone2018superpoint,lindenberger2023lightglue,chen2022aspanformer,sarlin2020superglue} can detect abundant feature points in challenging environments.
Methods\cite{chen2024camav2,mei2024rome,sarlin2023pixel} all utilized learning-based correspondence search methods.

In incremental reconstruction, efforts to improve the refinement step have been key to enhancing effectiveness. 
In the classic COLMAP system \cite{schonberger2016structure}, bundle adjustment (BA) optimizes camera poses, intrinsic parameters, and the positions of scene points by minimizing reprojection error. 
Weber \etal \cite{Weber_2023_CVPR} connected the bundle adjustment problem to power series theory and applied inverse expansion methods to achieve efficient large-scale BA. 
Zheng \etal \cite{Zheng_2023_ICCV} developed a distributed bundle adjustment approach using the exact Levenberg-Marquardt algorithm for extremely large datasets. 
Lindenberger \etal \cite{sarlin2023pixel} incorporated learned image features as optimization variables, eliminating errors from feature extraction which is difficult to address in traditional BA. 
% 写和上面不同的东西，我们在BA收到什么启发
We add rigid multi-camera set constraints to BA, ensuring that the optimized camera poses maintain a consistent relative relationship.
This can accelerate the convergence of BA and enhance its robustness.

\vspace{-0.1cm}
\subsection{Driving Scene Reconstruction and Aggregation}
\vspace{-0.1cm}
In practical engineering, driving data is often collected in a fragmented manner.
For driving scenes, MCSfM \cite{cui2023mcsfm} realizes scale-free driving scene reconstruction based on multi-camera systems. Aziza \etal \cite{zhanabatyrova2024structure} evaluate and compare the applicability and limitations of open-source SfM algorithms for mapping urban scenes.
For aggregating fragmented data, Chen \etal \cite{chen2020graph} propose a graph-based scene merging algorithm, which constructs a minimum spanning tree to find accurate similarity transformations and a minimum height tree to avoid error accumulation. 
Merge-SfM \cite{Fang2019MergeSfMMP} solves the problem of finding overlapping regions and the 7-DOF transformation between multiple reconstructions.
We first associate nearby scenes.
Then, the scene assembly binds them together in a coarse-to-fine fashion.

\vspace{-0.1cm}
\section{Methodology}
\vspace{-0.1cm}
\subsection{Problem Setting}
\vspace{-0.1cm}
Given a set of RGB images $I_1,\ldots,I_N$, where $N$ is the number of images and $I_i\in\mathbb{R}^{3\times{H}\times{D}}$, SfM estimates their corresponding intrinsic parameters $\mathbf{K}_i$, camera poses$\{\mathbf{R}_i, \mathbf{t}_i\}$ and the 3D scene represented by a point cloud$\{\mathbf{X}_1,\ldots, \mathbf{X}_M\}$, where $M$ is the number of scene points. 
3D scene points $\mathbf{X}_{j}$ can be projected as 2D points $\mathbf{x}_{ij}^{\prime}$ onto the pixel plane of image $I_i$:
\vspace{-0.2cm}
\begin{equation}\label{eq:project}
\mathbf{x}_{ij}^{\prime}=\pi(\mathbf{K}_{i},\mathbf{R}_{i},\mathbf{t}_{i},\mathbf{X}_{j}),
\vspace{-0.3cm}
\end{equation}
where the notation  $\pi\left(\cdot\right)$ is the projection function.

In MRASfM, multiple images captured at the same time construct a rigid unit. 
The pose of a rigid unit is defined as the pose of the vehicle at that moment. 
Let $U_{I_i}$ be the rigid unit that image $I_i$ belongs to, \ie, $I_i\in U_{I_i}$, and $\{\mathbf{R}_{U_{I_i}}, \mathbf{t}_{U_{I_i}}\}$ be the pose of rigid unit $U_{I_i}$. 
Let $\{\mathbf{R}_i^{rel}, \mathbf{t}_i^{rel}\}$ be the internal relative pose from image $I_i$ to its rigid unit $U_{I_i}$.  
The relationship between the camera pose of image $I_i$ and the pose of its corresponding rigid unit $U_{I_i}$ is as follows:
\vspace{-0.2cm}
 \begin{equation}
 \label{eq:ref_r}
     \mathbf{R}_{i}^{rel}=\mathbf{R}_{i}{\mathbf{R}_{U_{I_i}}}^{\mathrm{T}},
     \vspace{-0.2cm}
 \end{equation}
 \vspace{-0.4cm}
 \begin{equation}
 \label{eq:ref_t}
     \mathbf{t}_{i}^{rel}=\mathbf{R}_{U_{I_i}}(\mathbf{t}_{i}-\mathbf{t}_{U_{I_i}}).
     \vspace{-0.2cm}
 \end{equation}

\vspace{-0.1cm}
\subsection{System Framework}
\vspace{-0.1cm}
The pipeline of our SfM system is illustrated in Fig. \ref{fig:PIPELINE}. 
MRASfM takes multi-camera images, semantic information, rough trajectories and calibrations as input.  

In single-scene reconstruction, MRASfM first performs multi-camera correspondence search, which involves feature extraction, feature matching, and geometric validation. 
%By utilizing prior information, MRASfM is able to select suitable images for feature matching. 
Then, incremental reconstruction begins with prior-based initialization, reconstructing images from selected rigid units with the aid of prior information. 
Subsequently, MRASfM iteratively integrates each rigid unit into the reconstruction.
Each iteration step includes \textbf{(a)} Camera Set Registration: MRASfM robustly registers new images using rigid unit constraints; \textbf{(b)} Semantic-aided Triangulation: MRASfM triangulates scene points and removes outliers in the road surface area; \textbf{(c)} Camera Set BA: MRASfM refines the reconstruction with rigid units as optimization variables.

In multi-scene aggregation, MRASfM first associates scenes together using GNSS locations and selects two suitable scenes for aggregation.
They are coarsely assembled using reconstruction results.
%Then, images with large visual overlap are identified for feature matching and refinement of aggregation.
Then, an SfM-based optimization refines the transformation matrix between candidate scenes iteratively.
The fine assembly result serves as the new reference scene for the next iteration.
The process will repeat until all scenes are integrated.
%这个过程会循环进行，直到所有厂家都被融合进去

\vspace{-0.1cm}
\subsection{Correspondence Search and Initialization}
\label{subsec:Correspondence Search and Initialization}
\vspace{-0.1cm}
In correspondence search, feature points are detected by the Superpoint \cite{detone2018superpoint} model.
%because of its better performance in the weakly-textured areas \cite{chen2024camav2}. 
%Feature points corresponding to dynamic objects are filtered using semantic information.
With the known rough internal relative poses and vehicle poses, MRASfM calculates the approximate field of view for each image. 
Then, images with more visual overlap are selected for matching. 
%Specifically, for a given image $I_i$, MRASfM first identifies the rigid unit $u_j$ it belongs to and its nearest few units.
%Images from these units are selected as the candidates for matching. 
%If image $I_i$ and candidate images have similar orientations, feature matching will be performed between them. 
By filtering image pairs, both the efficiency and accuracy of the correspondence search are improved.
The Superglue model \cite{sarlin2020superglue} is used for feature matching, and matched point pairs of different semantic categories are removed.
Outliers of matching are then filtered via geometric verification \cite{hartley2003multiple}. 

After the multi-camera correspondence search, initial rigid units used for reconstruction are selected based on the total number of correspondences within the unit.
MRASfM directly registers images within initial rigid units using prior poses.
Then, the initial model is constructed through triangulation and BA.

\subsection{Camera Set Registration}\label{subsec:RIG_Registratio}
\begin{figure}[t]  
  \centering  
  \includegraphics[width=0.9\linewidth]{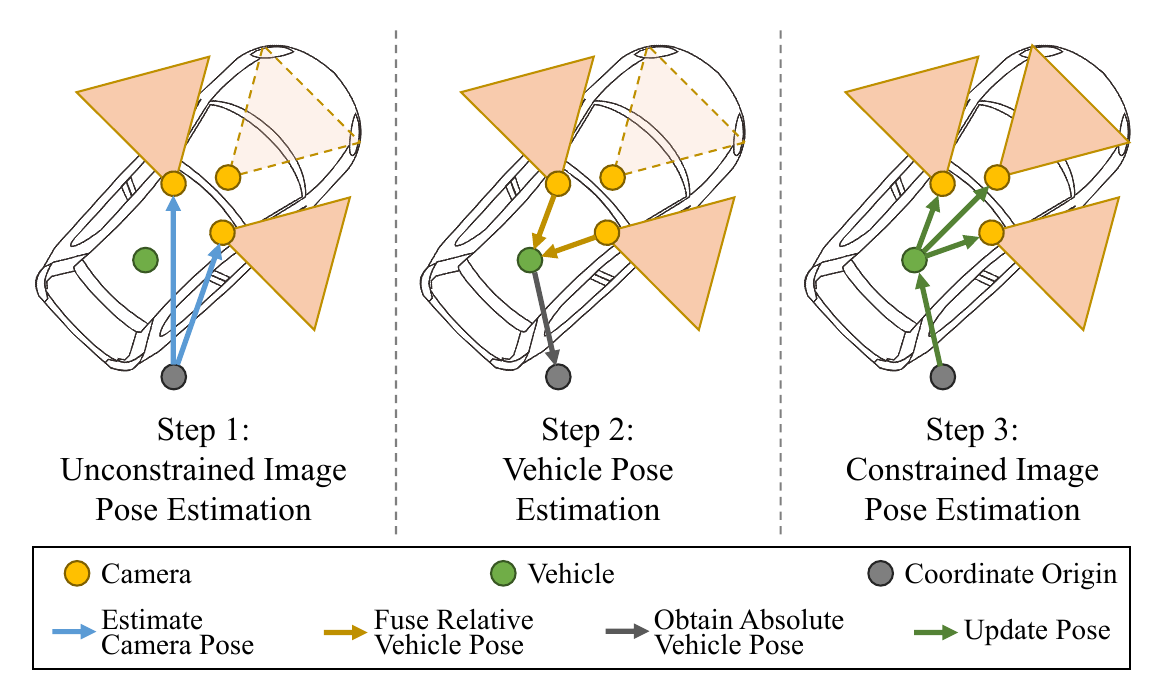} % 确保取消注释并使用实际图形文件  
  \vspace{-0.2cm}
  \caption{\textbf{The pipeline of Camera Set Registration:} camera pose estimation via PnP solving, multi-view fused vehicle pose derivation, and camera pose updating using intra-unit constraints and vehicle poses.}  
  \label{fig:rigregistration}  
  \vspace{-0.6cm}
\end{figure}  
Despite the refinement in correspondence search, occlusions in driving scenes can still lead to errors in pose estimation.
Camera set registration combines rigid constraints from camera sets and reliable estimation from correspondence-rich images to address this issue, as shown in Fig. \ref{fig:rigregistration}.

In unconstrained image pose estimation, MRASfM first organizes images using the next best view selection module in COLMAP \cite{schonberger2016structure}.
%A Perspective-n-Point (PnP) problem \cite{fischler1981random} is then formulated using feature correspondences to triangulated scene points.
The camera poses of the highest-ranked images are determined by solving the Perspective-n-Point (PnP) problem \cite{fischler1981random} using feature correspondences.
Let $\{\mathbf{R}_{k},\mathbf{t}_k\}$ ($k=1,\ldots,K$) be the newly estimated camera poses of image $I_k$, where $K$ is the number of estimations. 
According to Eq. \ref{eq:ref_r} and Eq. \ref{eq:ref_t}, these estimations can be used to calculate the poses of the rigid units where the aforementioned images belong:
\vspace{-0.3cm}
\begin{equation}
\begin{aligned}
    \mathbf{R}_{U_{I_k}}={\mathbf{R}_{k}^{rel}}^{\mathrm{T}}\mathbf{R}_{k},
\end{aligned}
\end{equation}
\vspace{-0.6cm}
\begin{equation}
    \mathbf{t}_{U_{I_k}}=\mathbf{t}_{k}-{\mathbf{R}_{k}^{rel}}^{\mathrm{T}}\mathbf{t}_{k}^{rel}.
    \vspace{-0.2cm}
\end{equation}
In vehicle pose estimation, for each rigid unit that has newly calibrated images, MRASfM uses the internal relative poses to infer its pose. 
For rigid unit $U_j$, the rotation of $U_j$ is computed by the local rotation averaging:
\vspace{-0.2cm}
\begin{equation}\label{get_rig_r}
    \mathbf{R}_{U_j}=\min_{\mathbf{R}_{U_j}}\sum\rho_{r}(\|\mathbf{R}_{U_j}\mathbf{R}_{U_{I_k}}\|),I_k\in U_j,
    \vspace{-0.4cm}
\end{equation}
where the notation $\rho\left(\cdot\right)$ is a robust loss function.
The  translation of $U_j$ is computed by the local translation averaging:
\vspace{-0.4cm}
\begin{equation}\label{get_rig_t}
    \mathbf{t}_{U_j}=\min_{\mathbf{t}_{U_j}}\sum\rho_{t}
    (\|\mathbf{t}_{U_j}-\mathbf{t}_{U_{I_k}}\|),I_k\in U_j,
    \vspace{-0.2cm}
\end{equation}

In constrained image pose estimation, the camera poses of all images belonging to $U_j$ can be obtained through the following formula:
\vspace{-0.3cm}
 \begin{equation}
 \begin{aligned}
\mathbf{R}_{i}=\mathbf{R}_{i}^{rel}\mathbf{R}_{U_j},I_i\in U_j,
\\
\mathbf{t}_{i}={\mathbf{R}_{U_j}}^{\mathrm{T}}\mathbf{t}_{i}^{rel}+\mathbf{t}_{U_j},I_i\in U_j.
 \end{aligned}
\vspace{-0.2cm}
 \end{equation}
where image $I_i$ belongs to the rigid unit $U_j$. 

In this way, although some images are difficult to register in driving scenes, MRASfM can still robustly estimate their camera poses in camera set registration.

\vspace{-0.1cm}
\subsection{Semantic-aided Triangulation}
\label{subsec:Semantic-assisted Triangulation}
\vspace{-0.1cm}

After registering a new rigid unit, MRASfM triangulates 3D points but faces challenges with dynamic feature points from vehicle shadows and insufficient road texture, leading to significant triangulation noise for the road surface. 
To address this, the Locally Optimized Random Sample Consensus (LO-RANSAC) method is applied to filter outliers by fitting a plane model to the 3D road points, thereby improving road surface reconstruction quality.
%After a new rigid unit is registered, MRASfM robustly and efficiently triangulates new 3D points. %new 2D-3D relationships can be established by triangulation. 
%However, the edges of vehicle shadows on the road surface can be detected as feature points. 
%When the vehicle moves, the feature points associated with its shadow are also dynamic. % 改 dynamic points
%These dynamic points cannot be filtered out by semantic information, leading to significant triangulation noise for many road points.
% 然而在阳光下，物体在路面上影子的边缘会被检测出特征点对。而当物体移动时，其影子所关联的特征点对也是动态的。这种动态点不能被语义信息所排除掉，从而引起很多地面点的三角化噪声。
%Moreover, the lack of texture on the road also results in many bad correspondences.
%Therefore, the triangulated road points typically contain a significant number of outliers. 
%To mitigate this issue, Locally Optimized Random Sample Consensus (LO-RANSAC) \cite{chum2003locally} is conducted to filter out outliers.
%MRASfM acquires the 3D road points with semantic information and estimates the best plane model using these points.
%Outliers are filtered based on the model, enhancing the quality of road surface reconstruction.
\vspace{-0.1cm}
\subsection{Camera Set BA}
\label{subsec:RIG BA}
\vspace{-0.1cm}
  \begin{figure}[t]  
  \centering  
  \includegraphics[width=0.85\linewidth]{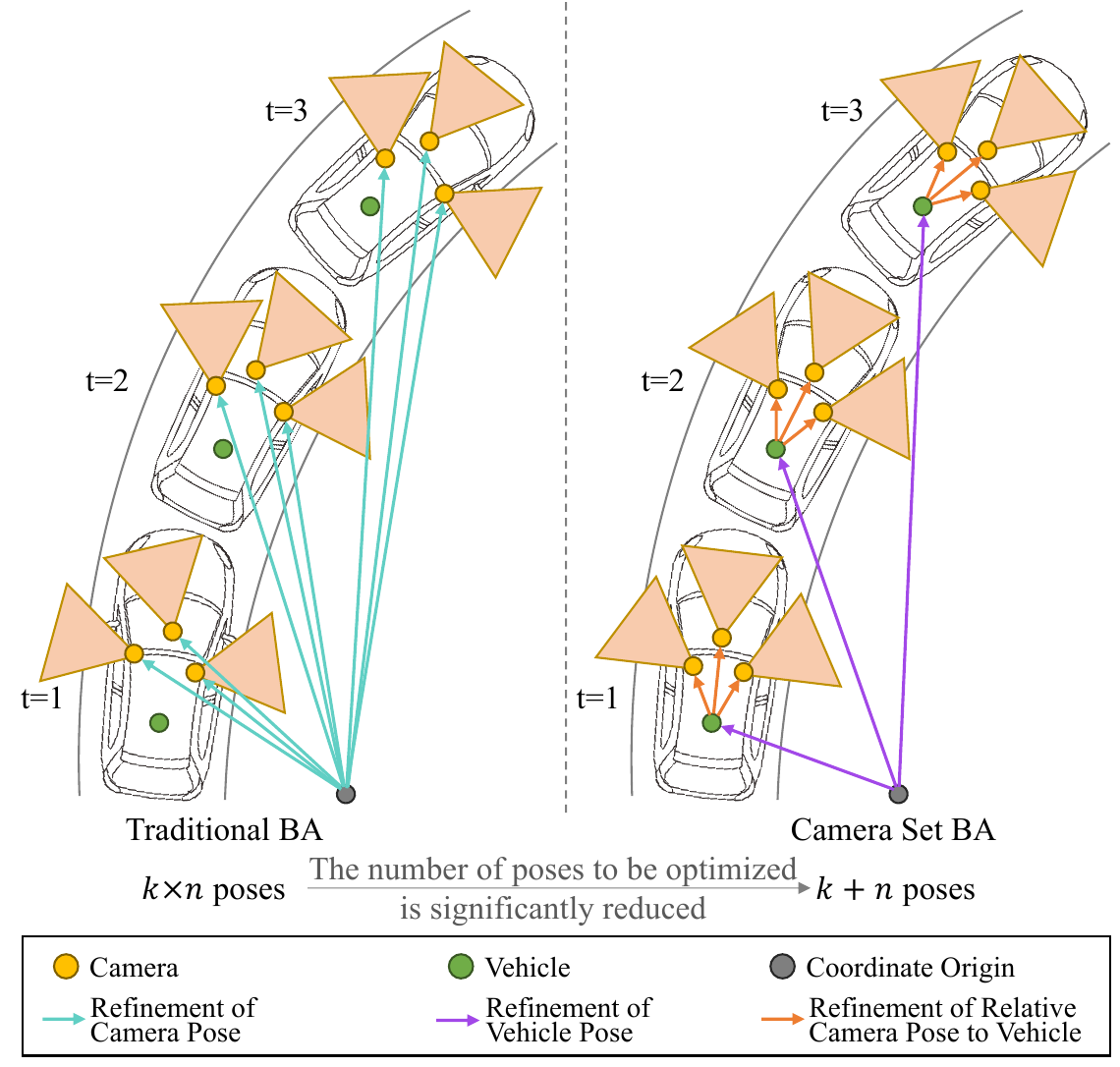} % 确保取消注释并使用实际图形文件  
  \vspace{-0.2cm}
  \caption{\textbf{Comparison between traditional BA and Camera Set BA.} 
  Traditional BA optimizes the absolute camera pose of each image individually. Camera Set BA, by contrast, optimizes vehicle poses and internal relative poses, significantly reducing optimization variables and enhancing robustness.}  
  \label{fig:RIGBA}  
  \vspace{-0.6cm}
\end{figure} 

In incremental SfM, BA optimizes registration and triangulation results.
Traditional BA optimizes each image individually, which can lead to inconsistent internal relative poses across frames. 
To address this, we introduce the Camera Set Bundle Adjustment (CSBA) module to enhance consistency. 
CSBA treats the camera set as a unit, optimizing vehicle poses and internal relative poses. 
Assuming that the multi-camera system comprises $k$ cameras and images from $n$ different timestamps are integrated into BA, traditional BA optimizes $k\times n$  poses, while CSBA only requires $k+n$. 
As the number of frames increases, CSBA significantly reduces the number of optimizing poses compared to traditional BA, conserving computational resources. 
The comparison between traditional BA and CSBA is shown in Fig. \ref{fig:RIGBA}.

To enhance system efficiency, the CSBA module is divided into Local CSBA and Global CSBA.
The system activates Global CSBA when there are a large number of unoptimized rigid units; otherwise, it executes Local CSBA.
The key difference is their optimization scope: Local CSBA optimizes only newly registered and their connected rigid units, while Global CSBA optimizes all registered units.
 \begin{figure}[t]  
  \centering  
  \includegraphics[width=0.9\linewidth]{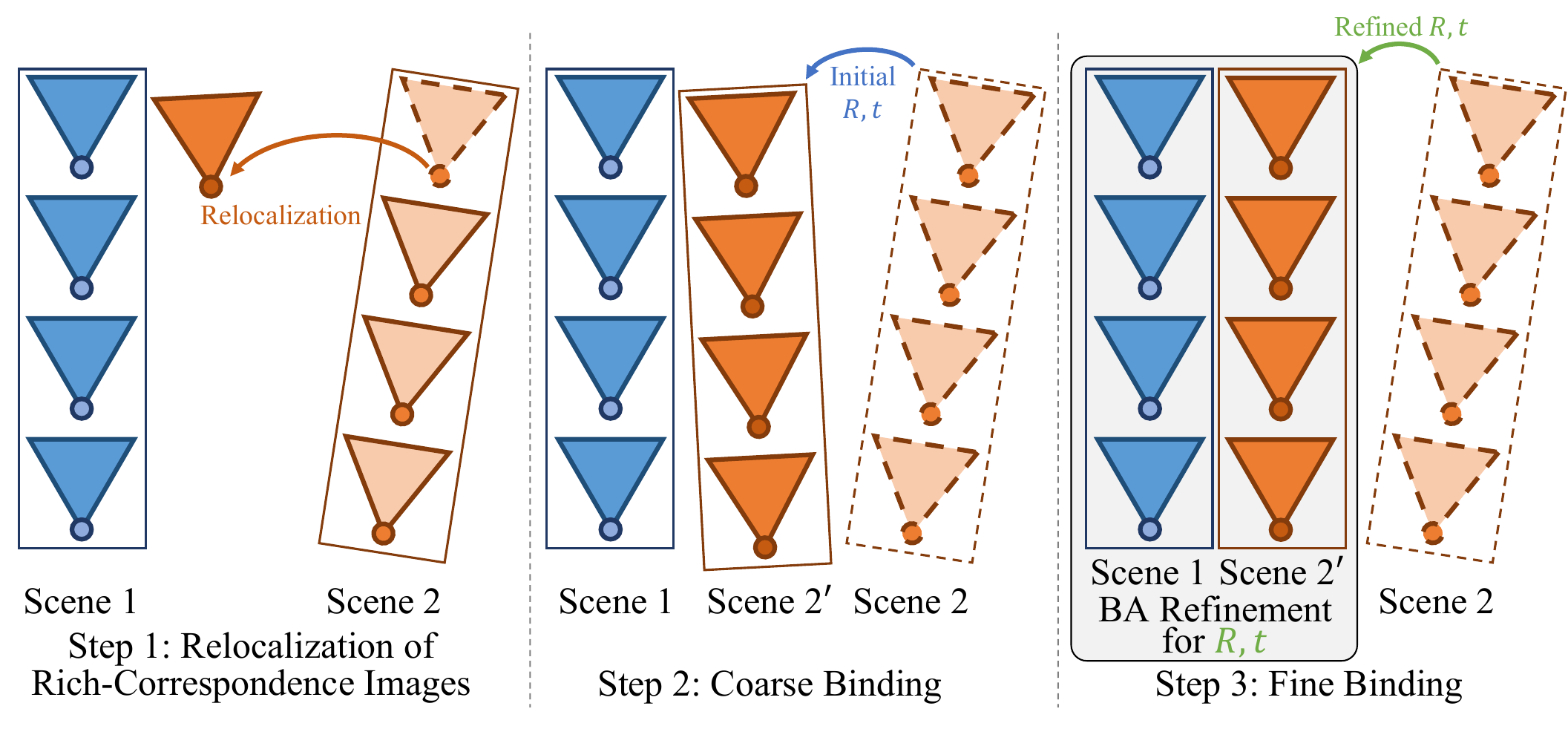} % 确保取消注释并使用实际图形文件  
  \vspace{-0.2cm}
  \caption{\textbf{Pipeline of Scene Assembly.} MRASfM begins by selecting images with rich correspondences for relocalization. By comparing the original and updated camera poses, the transformation matrix is initialized, which is then optimized by BA. The refined matrix is used to register remaining rigid units into reconstruction. Registration and refinement alternate until all rigid units are successfully integrated.
 }  
  \label{fig:Multi-scene Aggregation} 
  \vspace{-0.4cm}
\end{figure}  

According to \Cref{subsec:RIG_Registratio}, the camera pose of an image can be derived from its corresponding rigid unit pose and internal relative pose. 
Therefore, the parameters to be optimized include intrinsic camera parameters, vehicle poses, internal relative poses and scene points.
In local CSBA, intrinsic camera parameters and internal relative poses are fixed .
Let $\{\mathbf{R}_{l},\mathbf{t}_l\}$ ($l=1,\ldots,L$) be the camera poses of images integrated in local CSBA, which can be defined as:
\vspace{-0.2cm}
\begin{equation}
\begin{aligned}
\min_{\mathbf{R}_{U_{I_l}},\mathbf{t}_{U_{I_l}},\mathbf{X}_{j}}\sum\rho(\|\mathbf{x}_{lj}-\pi(\mathbf{K}_{l},\mathbf{R}_{l},\mathbf{t}_{l},\mathbf{X}_{j})\|^{2})
\\
\mathrm{s.t.}\quad\mathbf{R}_{l}=\mathbf{R}_{l}^{rel}\mathbf{R}_{U_{I_l}};\quad\mathbf{t}_{l}={\mathbf{R}_{U_{I_l}}}^{\mathrm{T}}\mathbf{t}_{l}^{rel}+\mathbf{t}_{U_{I_l}}.
\end{aligned}
\vspace{-0.2cm}
\end{equation}
where the notation $\pi(.)$ is the projection function in Eq. \ref{eq:project}, $\rho$(·) is the Cauchy loss function, $\mathbf{x}_{lj}$ are the 2D points in image $I_l$ corresponding to the 3D scene point $X_j$ and $K_{l}$ is the intrinsic parameter matrix of image $I_l$.

In Global CSBA, all aforementioned parameters are optimized.
Let $\{\mathbf{R}_{g},\mathbf{t}_g\}$ ($g=1,\ldots,G$) be poses of images to be optimized by global CSBA, which can be defined as:
\vspace{-0.2cm}
\begin{equation}
\begin{aligned}
\min_{\substack{\mathbf{K}_{g},\mathbf{R}_{U_{I_g}},\mathbf{t}_{U_{I_g}}, \\ \mathbf{R}_{g}^{rel},\mathbf{t}_{g}^{rel},\mathbf{X}_{j}}}\sum\rho(\|\mathbf{x}_{gj}-\pi(\mathbf{K}_{g},\mathbf{R}_{g},\mathbf{t}_{g},\mathbf{X}_{j})\|^{2})
\\
\mathrm{s.t.}\quad\mathbf{R}_{g}=\mathbf{R}_{g}^{rel}\mathbf{R}_{U_{I_g}};\quad\mathbf{t}_{g}={\mathbf{R}_{U_{I_g}}}^{\mathrm{T}}\mathbf{t}_{g}^{rel}+\mathbf{t}_{U_{I_g}}.
\end{aligned}
\vspace{-0.2cm}
\end{equation}
%For all the bundle adjustment modules, including LRBA and GRBA, the robust loss function $\rho$(·) is the Cauchy loss function.放到前面解释
By combining local CSBA and global CSBA, MRASfM can efficiently and robustly optimize the reconstruction process.

\vspace{-0.1cm}
\subsection{Multi-scene Aggregation}
\label{subsec:Multi-scene Aggregation}
\vspace{-0.1cm}

 \begin{figure*}[t]  
  \centering  
  \includegraphics[width=0.99\textwidth]{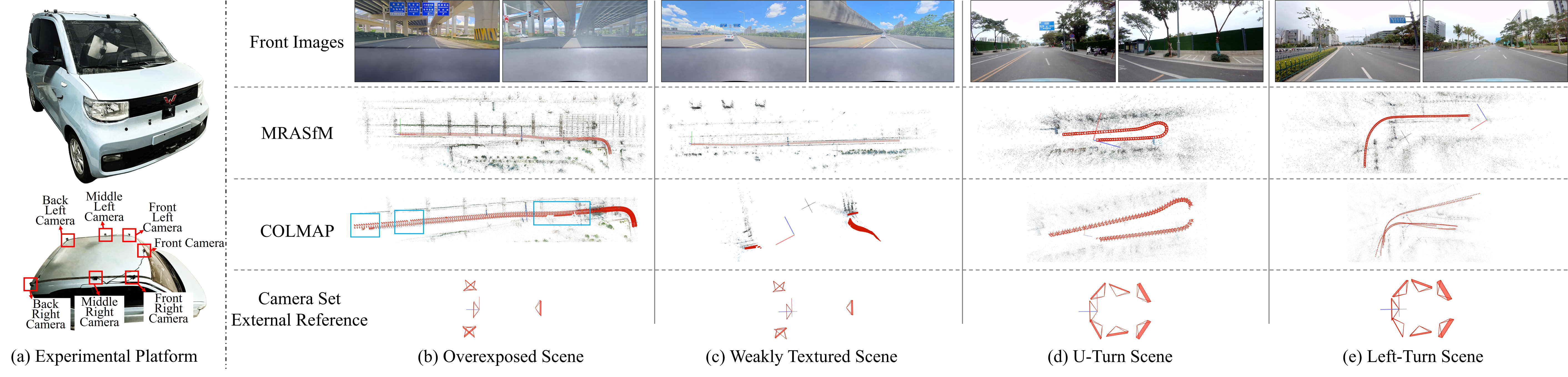} % 确保取消注释并使用实际图形文件  
  \vspace{-0.1cm}
  \caption{\textbf{Reconstruction results of self-collected datasets.} The calibrated internal relative poses of our system are shown in the last row. The camera poses are shown in red.
 }  
  \label{fig:self}  
  \vspace{-0.1cm}
\end{figure*}  
In practical engineering, scenes are frequently collected in a fragmented manner \cite{mei2024rome,zhang2024vision,wu2024emie}. 
Each scene typically contains only a short segment of driving information, and there are no shared images between different scenes. 
Additionally, associated scenes may be collected at long intervals.
To create a comprehensive and complete map, it is essential to seamlessly integrate multiple nearby scenes. 

In the scene association module, MRASfM first employs GNSS to position reconstructed scenes within a global coordinate system.
The locations of these scenes are represented by the midpoints of their trajectories and the scene located at the geometric center is considered the reference scene $S_r$. 
The scenes closest to the reference scene $S_r$ (typically three in experiments) are identified as candidate scenes for aggregation. 
Using homography-guided spatial pairs (HSP) from CAMAv2 \cite{chen2024camav2}, MRASfM assesses the visual overlap between two scenes and selects the candidate with the highest overlap as the merge scene $S_m$.

In the scene assembly module, shown in Fig. \ref{fig:Multi-scene Aggregation}, associated scenes $S_r$ and $S_m$ are first coarsely assembled within the global coordinate system.
The camera pose of image $I_i$ in the coarsely assembled scene $S_{fine}$ is defined as $\mathbf{P}_{i}^{coarse}$.
However, this assembly may differ from the actual situation by a transformation matrix $\mathbf{T}_{trans}$ due to GNSS errors.
Therefore, an iterative optimization for the transformation matrix $\mathbf{T}_{trans}$ is conducted for fine assembly.
To reduce computation time and memory usage, rigid units with large visual overlap are chosen for constructing SfM.
During optimization, MRASfM first initializes SfM by preserving the camera poses of the images from scene $S_r$ and triangulating them. 
Next, MRASfM selects the most suitable image $I_i$ from scene $S_m$ and performs unconstrained pose estimation (see Fig. \ref{fig:rigregistration}) to obtain its pose $\mathbf{P}_{i}^{fine}$ in the refined assembled scene $S_{fine}$. 
The initial transformation matrix $\mathbf{T}_{trans}$ can be obtained through the following formula:
\vspace{-0.2cm}
\begin{equation}
    \mathbf{T}_{trans} = \mathbf{P}_{i}^{fine}{\mathbf{P}_{i}^{coarse}}^{-1}
    \vspace{-0.2cm}
\end{equation}
where $\mathbf{P}_{i}^{coarse}$ is the camera pose of image $I_i$ from the coarsely assembled scene $S_{coarse}$. 

%By following the approach outlined in \Cref{subsec:RIG_Registratio}, 
Using transformation matrix $\mathbf{T}_{trans}$, we can register rigid unit $U_{I_i}$, which image $I_i$ belongs to:
\vspace{-0.1cm}
\begin{equation}
    \mathbf{P}_{U_{I_i}}^{fine} = \mathbf{T}_{trans}\mathbf{P}_{U_{I_i}}^{coarse}
    \vspace{-0.2cm}
\end{equation}
where $\mathbf{P}_{U_{I_i}}^{coarse}$ is the pose of rigid unit $U_{I_i}$ in scene $S_{coarse}$, and $\mathbf{P}_{U_{I_i}}^{fine}$ is the  pose of rigid unit $U_{I_i}$ in scene $S_{fine}$.
After triangulation outlined in \Cref{subsec:Semantic-assisted Triangulation},  
$\mathbf{T}_{trans}$ and the scene points will be optimized by transformation-based CSBA. 
Camera poses of images from the reference scene $S_r$ are fixed in BA.
\begin{figure}[t]  
  \centering  
  \includegraphics[width=0.9\linewidth]{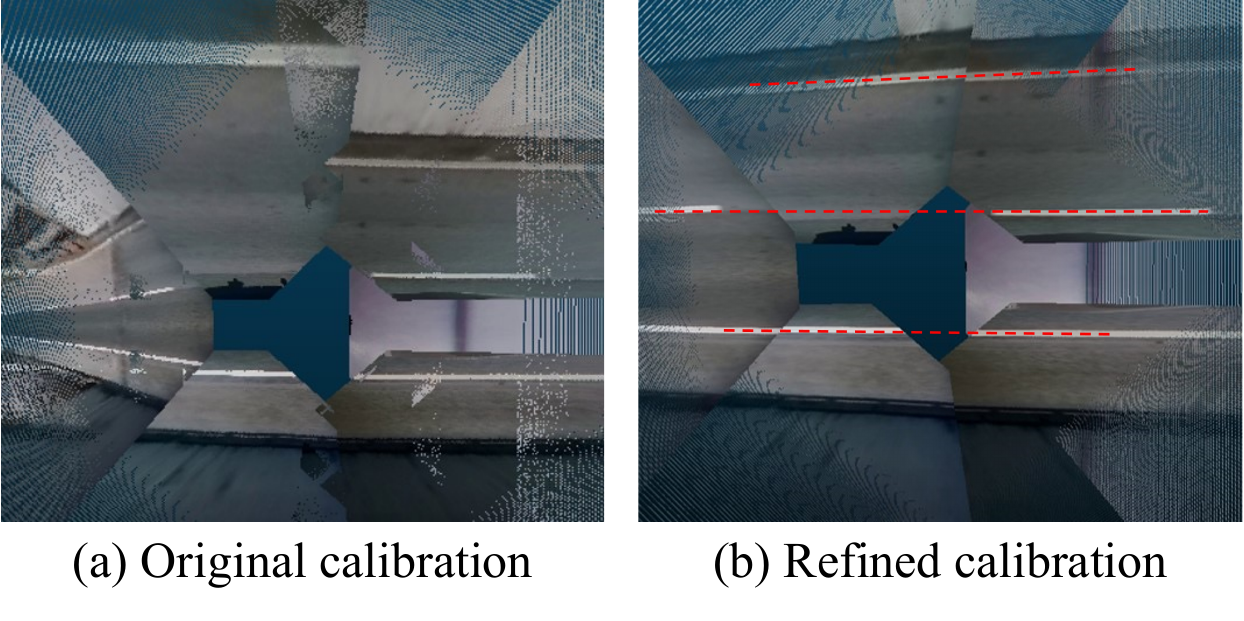} % 确保取消注释并使用实际图形文件  
  \vspace{-0.4cm}
  \caption{\textbf{BEV perspectives generated with calibrations.} The
refined calibrations generate more consistent line markings in BEV perspectives.
 }  
 \vspace{-0.2cm}
  \label{fig:extri}  
\end{figure}  

 \begin{figure}[t]  
  \centering  
  \includegraphics[width=0.99\linewidth]{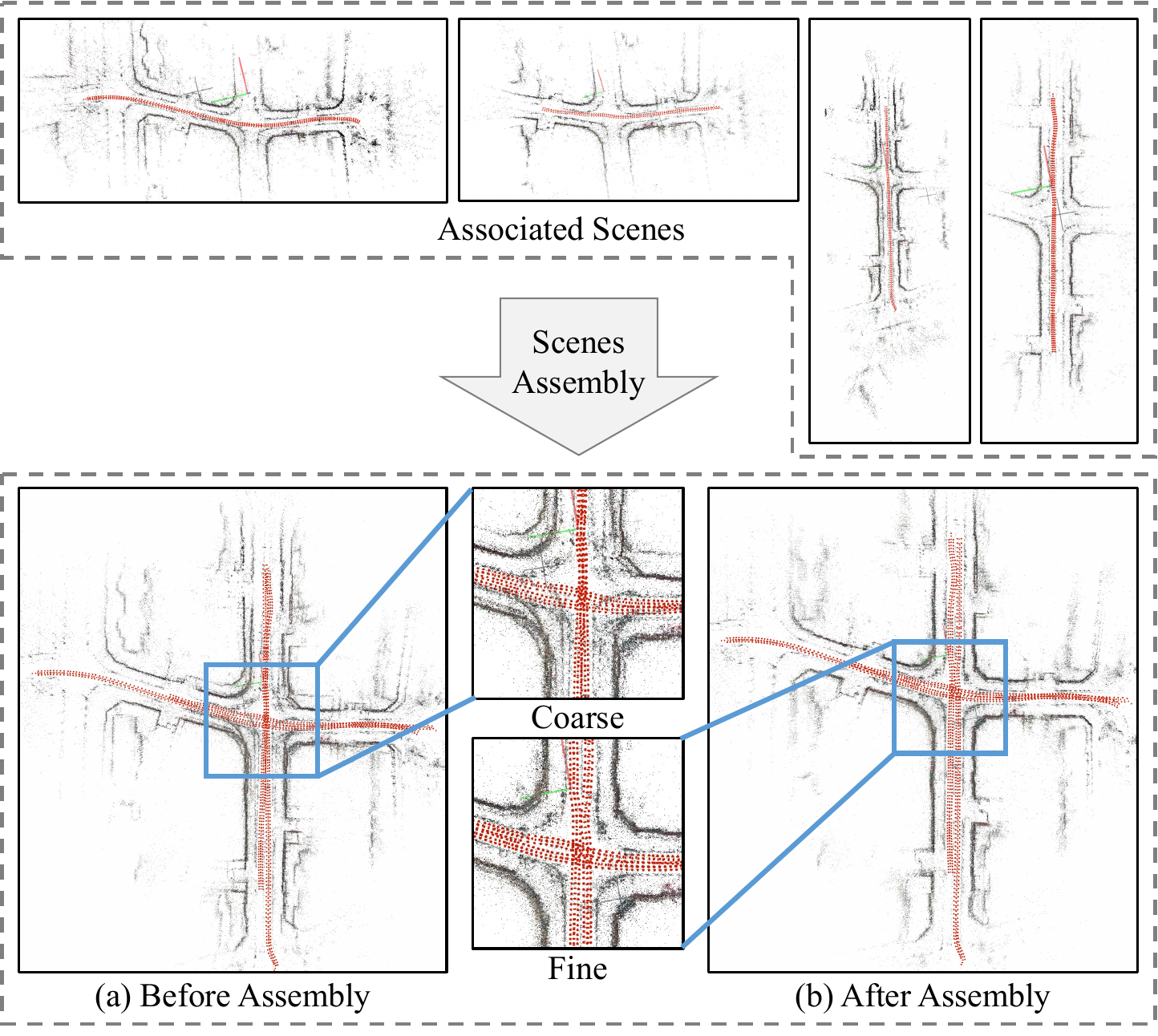} % 确保取消注释并使用实际图形文件  
  \vspace{-0.7cm}
  \caption{\textbf{Results of Multi-Scene Aggregation.} The associated scenes are integrated into a complete and consistent scene through the multi-scene aggregation module. 
 }  
  \label{fig:merge_result}  
  \vspace{-0.5cm}
\end{figure}  
Let $\{\mathbf{R}_{m}^{coarse},\mathbf{t}_{m}^{coarse}\}$ ($m=1,\ldots,M$) be the camera poses of registered images from scene $S_m$. 
The transformation-based BA can be defined as: 
\vspace{-0.1cm}
\begin{equation}
\begin{aligned}
\min_{\mathbf{T}_{trans},\mathbf{X}_{j}}\sum\rho(\|\mathbf{x}_{mj}-\pi(\mathbf{K}_{m},\mathbf{R}_{m},\mathbf{t}_{m},\mathbf{X}_{j})\|^{2})
\\
\text{s.t.} \quad \mathbf{R}_{m}=\mathbf{R}_{\text{trans}} \mathbf{R}_{m}^{\text{coarse}};
\mathbf{t}_{m}=\mathbf{R}_{\text{trans}} \mathbf{t}_{m}^{\text{coarse}} + \mathbf{t}_{\text{trans}}.  
\end{aligned}
\vspace{-0.2cm}
\end{equation}
where $\mathbf{R}_{trans}$ and $\mathbf{t}_{trans}$ are the rotation matrix and the translation vector decomposed from $\mathbf{T}_{trans}$. 
Camera set registration, triangulation, and transformation-based BA will iteratively take place until all rigid units in scene $S_m$ are integrated into scene $S_{fine}$. 
Thereby scene $S_r$ and scene $S_m$ are seamlessly aggregated as scene $S_{fine}$, which serves as the new reference scene for the remaining ones.
In a circular manner, all scenes are integrated into a complete map through scene association and scene assembly.

\vspace{-0.1cm}
\section{Real-World Applications}
\vspace{-0.1cm}
We evaluate MRASfM using datasets from two multi-camera systems: a six-camera and a seven-camera surround-view setup (see Fig. \ref{fig:self} (a)).
These experimental platforms operate at speeds of 10 to 60 km/h, capturing high-resolution (1920 × 1080) surround-view imagery at 30 Hz, with simultaneous GNSS data recorded.  
All the experiments were conducted on a computer equipped with a 3.4 GHz CPU.

Sample reconstructed scenes and camera poses are shown in Fig. \ref{fig:self}. 
Scene (b) and (c) were captured with a six-camera system in an urban driving environment characterized by weak textures and slope changes, and were successfully reconstructed by MRASfM.
Scene (d) and (e), captured with a seven-camera system, further demonstrate MRASfM’s robustness in complex U‑turn and left‑turn scenarios.
Compared to COLMAP \cite{schonberger2016structure}, MRASfM shows superior generalizability and robustness under challenging conditions.
This improved performance is primarily attributed to accurate correspondence search and effective integration of rigid multi-camera constraints.

A key feature of MRASfM is its ability to mitigate calibration inaccuracies.
Due to the difficulty of vehicle calibration, initial calibration inherently contains errors, which can worsen when driving on uneven terrain.
The bird's-eye views (BEV) in Fig. \ref{fig:extri}, generated using calibration data, demonstrate improved lane marking alignment, visually confirming MRASfM's recalibration capability. 
%The bird's eye views (BEV) perspectives are generated using calibration data in Fig. \ref{fig:extri}.
%The improved consistency of lane marking alignment in the BEV images visually validates the recalibration capability of MRASfM . 
By treating rigid camera sets as fundamental units within BA, MRASfM effectively corrects initial calibration errors, yielding a more accurate and consistent multi-camera system calibration.

For scenes collected in a fragmented manner, the Multi-Scene Aggregation module effectively integrates them into a cohesive whole.
Fig. \ref{fig:merge_result} illustrates the difference between coarse binding and fine binding.
Due to initial GNSS errors, fragmented scenes cannot form a coherent map directly. 
However, after the iterative optimization of transformation matrices, a complete and consistent map is finally achieved.
\begin{table*}[]\footnotesize 
\centering
\setlength{\tabcolsep}{1.6mm}
\renewcommand\arraystretch{1.0}
\caption{\textbf{Vehicle pose accuracy of KITTI odometry benchmark.}
$e_r$ is the median absolute rotation error (degrees);
$e_t$ is the median absolute translation error (meters);
$T$ is running time of pose estimation (minutes).
The best results are shown in \textbf{bold}; the second best are \underline{underlined}.}
\vspace{-0.2cm}
\begin{tabular}{c|cc|cc|cc|ccc|ccc}
\hline
Data & \multicolumn{2}{c|}{COLMAP\cite{schonberger2016structure}} & \multicolumn{2}{c|}{GLOMAP\cite{pan2024global}} & \multicolumn{2}{c|}{MGSfM\cite{tao2025mgsfm}} & \multicolumn{3}{c|}{MCSfM\cite{cui2023mcsfm}} & \multicolumn{3}{c}{MRASfM} \\ \hline
Name   & $e_r$ ↓                         & $e_t$ ↓                         & $e_r$ ↓        & $e_t$ ↓    & $e_r$ ↓      & $e_t$ ↓      & $e_r$ ↓                & $e_t$ ↓               & $T$ ↓          & $e_r$ ↓      & $e_t$ ↓      & $T$ ↓ \\ \hline
data00 & \underline{0.4} & 0.9 & \underline{0.4} & 0.8 & 0.4 & \underline{0.5} & \textbf{0.3} & \underline{0.5} & 286 & 0.5 & \textbf{0.3} & 192 \\
data01 & \textbf{0.2} & 1.5 & 0.5 & 4.5 & \textbf{0.2} & \textbf{0.6} & \underline{0.4} & \underline{1.0} & 47 & \textbf{0.2} & \textbf{0.6} & 34 \\
data02 & 0.5 & 4.0 & \underline{0.4} & 5.4 & \underline{0.4} & \underline{0.9} & \textbf{0.3} & 1.0 & 355 & \underline{0.4} & \textbf{0.7} & 276 \\
data03 & \textbf{0.1} & \underline{0.2} & \underline{0.2} & 0.4 & \underline{0.2} & \underline{0.2} & \underline{0.2} & \underline{0.2} & 77 & \underline{0.2} & \textbf{0.1} & 48 \\
data04 & \textbf{0.1} & \textbf{0.1} & \textbf{0.1} & \underline{0.2} & \textbf{0.1} & \textbf{0.1} & \textbf{0.1} & \textbf{0.1} & 4 & \textbf{0.1} & \textbf{0.1} & 3 \\
data05 & 0.6 & 0.8 & \textbf{0.2} & 0.3 & \textbf{0.2} & \underline{0.2} & \underline{0.3} & \underline{0.2} & 116 & \underline{0.3} & \textbf{0.1} & 81 \\
data06 & \underline{0.2} & \underline{0.2} & \textbf{0.1} & 0.3 & \textbf{0.1} & \textbf{0.1} & \textbf{0.1} & \underline{0.2} & 34 & \textbf{0.1} & \textbf{0.1} & 28 \\
data07 & 1.1 & 0.5 & 0.3 & \underline{0.3} & \underline{0.2} & \textbf{0.2} & 0.4 & 0.4 & 62 & \textbf{0.1} & \textbf{0.2} & 43 \\
data08 & 0.7 & 6.1 & 0.7 & 3.1 & \underline{0.4} & \underline{0.9} & \underline{0.4} & 1.2 & 276 & \textbf{0.3} & \textbf{0.5} & 188 \\
data09 & 0.6 & 1.1 & \textbf{0.3} & 1.7 & \textbf{0.3} & \underline{0.5} & \textbf{0.3} & \underline{0.5} & 74 & \textbf{0.3} & \textbf{0.2} & 54 \\
data10 & 0.8 & 2.5 & \textbf{0.3} & 0.7 & \underline{0.4} & 0.6 & \underline{0.4} & \underline{0.4} & 53 & \underline{0.4} & \textbf{0.2} & 38 \\ \hline
\end{tabular}
\vspace{-0.2cm}
\label{table:odometry}
\end{table*}

 \begin{figure*}[t]  
  \centering  
  \includegraphics[width=0.99\linewidth]{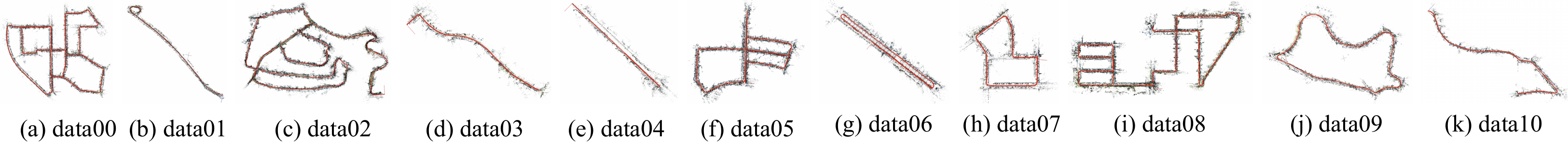} % 确保取消注释并使用实际图形文件  
  \vspace{-0.3cm}
  \caption{\textbf{Reconstruction results of KITTI odometry benchmark produced by MRASfM.}
 }  
  \label{fig:odo}  
  \vspace{-0.6cm}
\end{figure*}  

\vspace{-0.1cm}
\section{Large-Scale Validation Results}
\vspace{-0.1cm}
%The stereo camera dataset comes from the KITTI odometry benchmark \cite{geiger2012we}, and one six-camera surrounding systems comes from the nuScenes \cite{nuscenes} dataset, where the ground-truth camera poses are provided in both cases. 
%The four-camera dataset comes from KITTI-360 benchmark \cite{liao2022kitti}, which offers some ground-truth camera poses.  
%One six-camera surrounding system dataset and the seven-camera surrounding system are self-collected with rough poses from GNSS.

\vspace{-0.05cm}
\subsection{KITTI odometry}
\vspace{-0.05cm}
The KITTI odometry benchmark dataset is collected using a car equipped with a stereo camera, encompassing 11 sequences of urban driving scenes. 
The quantitative experiment results are presented in TABLE \ref{table:odometry}. 
The median absolute pose error (APE) is calculated for evaluation.
Our method achieves state-of-the-art performance in terms of pose estimation accuracy.
This is mainly due to iterative refinement and effective utilization of prior information.
% 直接说我们的方法需要更少的时间
Meanwhile, MRASfM significantly outperforms the previous state-of-the-art incremental SfM method, MCSfM\cite{cui2023mcsfm}, in terms of reconstruction efficiency.
This improvement is largely attributed to our selection of high-quality matching image pairs during the correspondence search. 
% 每轮循环的重建都准确，加快了收敛
Moreover, the accurate registration and triangulation accelerate the convergence of refinement.
The qualitative experiment results shown in Fig. \ref{fig:odo} demonstrate that MRASfM is able to achieve consistent reconstruction across various scenes

%In non-looping sequences, such as data01 and data02, SLAM and global SfM methods struggle with accurate trajectory estimation, as indicated in TABLE \ref{table:odometry}. 
%For SLAM methods, the primary reason is the absence of loop closure detection, which hampers the mitigation of accumulated errors. 
%For global SfM, challenges arise mainly from collinear relative translations and feature match outliers. 
%In contrast, MRASfM repeatedly removes matching outliers through CSBA module and robustly registers images during camera set registration.
%In looping sequences, MRASfM still outperforms other methods.
%This is primarily owing to the multi-camera correspondence search module, which can perform feature matching between images of closure loop to alleviate cumulative errors.
%The qualitative experiment results shown in Fig. \ref{fig:odo} demonstrates that MRASfM is able to achieve consistent reconstruction across various scenes
\vspace{-0.1cm}
\subsection{NuScenes}
\begin{figure}[t]  
  \centering  
  \includegraphics[width=0.99\linewidth]{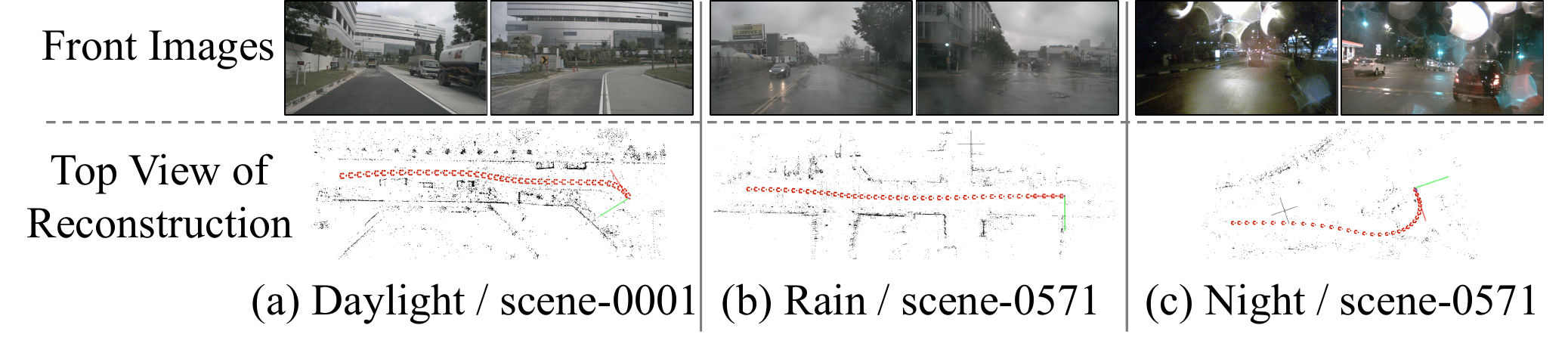} % 确保取消注释并使用实际图形文件  
  \vspace{-0.5cm}
  \caption{\textbf{Reconstruction results on the nuScenes dataset.}
 }  
  \label{fig:nuscenes}  
  \vspace{-0.6cm}
\end{figure}

\vspace{-0.1cm}
The nuScenes dataset is a public large-scale dataset for autonomous driving, which is collected using a vehicle equipped with six surrounding cameras.
The dataset covers over 1000 scenes, each lasting about
20 seconds, which spread across different countries, lighting
settings, weather variations, and environments.
The quantitative experiment results are presented in TABLE \ref{table:nuscenes}. 
Our method also achieves state-of-the-art performance in terms of pose estimation accuracy.
Furthermore, qualitative results in Fig. \ref{fig:nuscenes} showcase the capacity of MRASfM to consistently reconstruct scenes across the diverse and challenging environments within the nuScenes dataset. 
MRASfM demonstrates robust performance across both quantitative metrics and qualitative visual assessments.
These accurate and robust results are primarily attributed to the improved correspondence research and the effective utilization of rigid unit constraints during the reconstruction process.

\begin{table}[]\footnotesize 
\centering
\setlength{\tabcolsep}{4.5mm}
\renewcommand\arraystretch{1.0}
\caption{\textbf{Vehicle pose accuracy of nuScenes dataset.} 
$e_t$ is the RMSE absolute translation error in meters. 
The best results are shown in bold.}
\vspace{-0.2cm}
\begin{tabular}{c|c|c}
\hline
Method                                           & Input            & $e_t$ ↓        \\ \hline
ORBSLAM3 \cite{campos2021orb}   & Front camera     & 0.199          \\
DROID-SLAM \cite{teed2021droid} & Front camera     & 0.282          \\
OCC-VO \cite{li2024occ-vo}      & Surround cameras & 0.140          \\
GLOMAP\cite{pan2024global}      & Surround cameras & 0.158          \\
MGSfM\cite{tao2025mgsfm}        & Surround cameras & 0.134          \\
MRASfM                                           & Surround cameras & \textbf{0.124} \\ \hline
\end{tabular}
\vspace{-0.2cm}
\label{table:nuscenes}
\end{table}

\vspace{-0.2cm}
\subsection{Ablation Study}
\vspace{-0.2cm}
To validate each module of MRASfM, reconstructions of KITTI odometry benchmark data00 and data01 under different conditions are presented in TABLE \ref{table:ablation}.

\textbf{The effects of camera set BA} are demonstrated in TABLE \ref{table:ablation} (a). The absence of CSBA leads to a noticeable decline in both efficiency and accuracy, primarily due to the increased number of optimization variables.
Additionally, the lack of camera set constraints increases optimization errors and prolongs convergence times.

\textbf{The impact of camera set registration} is evaluated in TABLE \ref{table:ablation} (b).
While it has minimal effect on pose estimation accuracy and reconstruction efficiency, it is particularly useful for handling occluded viewpoints. 
In scenes with rich correspondences, its impact is less noticeable, though it still improves reconstruction by enhancing image registration.

\begin{figure}[t]  
  \centering  
  \includegraphics[width=0.99\linewidth]{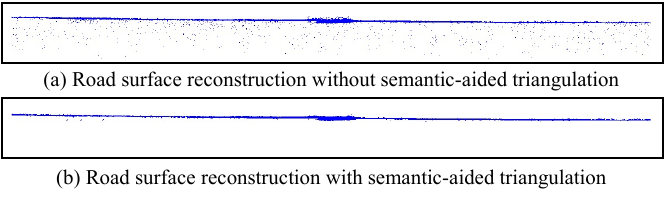} % 确保取消注释并使用实际图形文件  
  \vspace{-0.6cm}
  \caption{\textbf{Comparison between triangulated road points.} The semantic-aided triangulation module effectively filters outliers of the road surface.
 }  
  \label{fig:groundpoints} 
  \vspace{-0.6cm}
\end{figure}

\textbf{The influences of semantic-aided triangulation} are examined in TABLE \ref{table:ablation} (c).
Semantic-aided triangulation improves reconstruction accuracy and accelerates optimization convergence by removing outliers. Although its impact on pose estimation and efficiency is minimal, its effectiveness is evident in the enhanced road surface reconstruction shown in Fig. \ref{fig:groundpoints}.
\begin{table}[t]\scriptsize 
\centering
\caption{\textbf{Ablation study results of MRASfM on vehicle pose estimation.} }
\setlength{\tabcolsep}{0.3mm}
\vspace{-0.2cm}
\renewcommand\arraystretch{1.1}
\begin{tabular}{c|l|ccc|ccc}
\toprule
% \hline
\multirow{2}{*}{Exp.} & \multirow{2}{*}{Method}                      & \multicolumn{3}{c|}{data00}               & \multicolumn{3}{c}{data01}                \\ \cline{3-8} 
                      &                                              & $e_r$ ↓        & $e_t$ ↓        & $T$ ↓         & $e_r$ ↓        & $e_t$ ↓        & $T$ ↓         \\ \hline
\multirow{2}{*}{(a)}    & Ours (w/o camera set BA)                     & 1.8          & 2.7          & 8720        & 0.7          & 1.0          & 534         \\
 & Ours (full, w/ camera set BA)                & \textbf{0.5} & \textbf{0.3} & \textbf{192} & \textbf{0.2} & \textbf{0.6} & \textbf{34}     \\ \hline
\multirow{2}{*}{(b)}    & Ours (w/o camera set registration)           & 0.6          & 0.4          & 203          & 0.3          & 0.7          & 40          \\
                      & Ours (full, w/ camera set registration)              & \textbf{0.5} & \textbf{0.3} & \textbf{192} & \textbf{0.2} & \textbf{0.6} & \textbf{34} \\ \hline
\multirow{2}{*}{(c)}    & Ours (w/o semantic-aided triangulation)      & 0.6          & \textbf{0.3} & 197          & \textbf{0.2} & \textbf{0.6} & 37 \\
                      & Ours (full, w/ semantic-aided triangulation) & \textbf{0.5} & \textbf{0.3} & \textbf{192} & \textbf{0.2} & \textbf{0.6} & \textbf{34} \\ \bottomrule

\end{tabular}
\label{table:ablation}
\vspace{-0.2cm}	
\end{table}

% \vspace{-0.1cm}
\section{Conclusion}
% \vspace{-0.1cm}
In this work, we propose MRASfM, a novel multi-camera reconstruction framework for driving scene reconstruction.
Our framework uses camera sets as atomic units for registration and refinement, enabling robust pose estimation in complex environments and higher efficiency than conventional per-camera methods.
During triangulation, the quality of road surface reconstruction is greatly enhanced using semantic information. 
For fragmented scenes, the proposed multi-scene aggregation module seamlessly binds nearby scenes into a complete map using a coarse-to-fine approach.
In real-world applications, MRASfM demonstrates its generalizability and robustness through its superior performance compared to COLMAP.
Public datasets evaluations validate the state-of-the-art performance of MRASfM. 
MRASfM enables intelligent vehicles to better perceive environments and estimate ego poses, which is crucial for downstream tasks.

\normalem
\bibliographystyle{IEEEtran}  % set style to IEEE
\bibliography{IEEEabrv,root} % set reference file name

\end{document}